\documentclass[runningheads]{llncs}
\usepackage{graphicx}
\usepackage{amsmath}
\usepackage{bbm}
\usepackage{multirow}
\usepackage{bbding}
\usepackage{color}
\usepackage{booktabs}
\usepackage{url}

\begin{document}
\newcommand{\xp}[1]{{\color[rgb]{0.9,0.1,0.1}{[XP:#1]}}}
\newcommand{\xmli}[1]{{\color[rgb]{0.0,0.1,0.9}{[XM:#1]}}}
\newcommand{\zheang}[1]{{\color[rgb]{0.0,0.4,0.1}{[ZA:#1]}}}

\iffalse
\def\etal{\textit{et al}. }
\def\ie{\textit{i.e.}}
\def\eg{\textit{e.g.}}
\def\vs{\textit{v.s.}}
\fi

\def\etal{{et al}.}
\def\ie{{i.e.}}
\def\eg{{e.g.}}
\def\vs{{v.s.}}
\title{Context-Aware Pseudo-Label Refinement for Source-Free Domain Adaptive Fundus Image
Segmentation}
%\thanks{Supported by organization x.}###

\titlerunning{CPR-SFUDA}
%\author{Zheang Huai, Xinpeng Ding, Yi Li, and Xiaomeng Li}
%\institute{}

\author{Zheang Huai \and
Xinpeng Ding \and
Yi Li \and
Xiaomeng Li\thanks{Correspondence: {\tt\small eexmli@ust.hk}}
}
%\textsuperscript{(\Letter)}
% index{Huai, Zheang}
% index{Ding, Xinpeng}
% index{Li, Yi}
% index{Li, Xiaomeng}
%
\authorrunning{Z. Huai et al.}
\institute{The Hong Kong University of Science and Technology, Kowloon, Hong Kong 
%\email{eexmli@ust.hk}
}

\iffalse
\author{First Author\inst{1}\orcidID{0000-1111-2222-3333} \and
Second Author\inst{2,3}\orcidID{1111-2222-3333-4444} \and
Third Author\inst{3}\orcidID{2222--3333-4444-5555}}

%
\authorrunning{F. Author et al.}
\institute{Princeton University, Princeton NJ 08544, USA \and
Springer Heidelberg, Tiergartenstr. 17, 69121 Heidelberg, Germany
\email{lncs@springer.com}\\
\url{http://www.springer.com/gp/computer-science/lncs} \and
ABC Institute, Rupert-Karls-University Heidelberg, Heidelberg, Germany\\
\email{\{abc,lncs\}@uni-heidelberg.de}}
\fi
%
\maketitle              % typeset the header of the contribution
\begin{abstract}
In the domain adaptation problem, source data may be unavailable to the target client side due to privacy or intellectual property issues. Source-free unsupervised domain adaptation (SF-UDA) aims at adapting a model trained on the source side to align the target distribution with only the source model and unlabeled target data. The source model usually produces noisy and context-inconsistent pseudo-labels on the target domain, i.e., neighbouring regions that have a similar visual appearance are annotated with different pseudo-labels. 
This observation motivates us to refine pseudo-labels with context relations. 
%
%To this end, we propose a context-aware pseudo-label refinement method for SF-UDA to refine pseudo-labels with context relations.
%
%The key motivation is based on the fact that features of the same class tend to form a cluster despite the domain gap, which implies context relations can be readily calculated from feature distances.
%
%Our method consists of three modules: a context-similarity learning module, a pseudo-label refinement module and a denoising module.
%
%The context-similarity learning module aims to learn context relations by calculating the distances of features
Another observation is 
%the inherent feature distribution characteristic
that features of the same class tend to form a cluster despite the domain gap, which implies context relations can be readily calculated from feature distances. To this end, %\xmli{we propose a novel Context-Aware Pseudo-Label Refinemen method for SFUDA. The key motivation of our method is xxxx. Our method consists of three modules: xxx, xxx, xxxx. XXX aims to learn context x. -- goal xx. By using these together, we address problems. }
we propose a context-aware pseudo-label refinement method for SF-UDA. Specifically, a context-similarity learning module is developed to learn context relations. Next, pseudo-label revision is designed utilizing the learned context relations. Further, we propose calibrating the revised pseudo-labels to compensate for wrong revision caused by inaccurate context relations. Additionally, we adopt a pixel-level and class-level denoising scheme to select reliable pseudo-labels for domain adaptation. Experiments on cross-domain fundus images indicate that our approach yields the state-of-the-art results. Code is available at \textcolor{blue}{\url{https://github.com/xmed-lab/CPR}}.
%~\footnote{Our code will be released upon paper acceptance.}
\end{abstract}
\section{Introduction}
%\subsection{A Subsection Sample}
Accurate segmentation of the optic cup and optic disc in fundus images is essential for the cup-to-disc ratio measurement that is critical for glaucoma screening and detection~\cite{fundus_seg_background}. Although deep neural networks have achieved great advances in medical image segmentation, they are susceptible to data with domain shifts, such as those caused by using different scanning devices or different hospitals \cite{beal}. 
Unsupervised domain adaptation \cite{uda} is proposed to transfer knowledge to the target domain with access to the source and target data while not requiring any annotation in the target domain. 
Recently, source-free unsupervised domain adaptation (SF-UDA) has become a significant area of research~\cite{de,perturb,3cgan,shot,ug,prabhu}, where source data is inaccessible due to privacy or intellectual property concerns.

%\subsubsection{Related Work} 
Existing SF-UDA solutions can be  categorized into four main groups: batch normalization (BN) statistics adaptation \cite{memory,bnmiccai,tent}, approximating source images \cite{prompt,generation}, entropy minimization \cite{adaent}, and pseudo-labeling \cite{dpl,ud4r}. 
BN statistics adaptation methods aim to address the discrepancy of statistics between different domains. For example, \cite{memory,bnmiccai} update low-order and high-order BN statistics with distinct training objectives, while \cite{tent} adapts BN statistics to minimize the entropy of the model's prediction.
Approximating source images aims to generate source-like images. For example, \cite{generation} first attains a coarse source image by freezing the source model and training a learnable image, then refines the image via mutual Fourier Transform. 
The refined source-like image provides a representation of the source data distribution and facilitates domain alignment during the adaptation process. For another instance, \cite{prompt} learns a domain prompt to add to a target domain image so that the sum simulates the source image.
Entropy minimization methods aim to produce more confident model predictions. For example, \cite{adaent} minimizes output entropy with a regularizer of class-ratio. The class-ratio is estimated by an auxiliary network that is pre-trained on the source domain.
For pseudo-labeling \cite{pseudolabel,yao2022enhancing}, erroneous pseudo-labels are either discarded or corrected. For example, \cite{dpl} identifies low-confidence pseudo-labels at both the pixel-level and the class-level. On the other hand, \cite{ud4r} performs uncertainty-weighted soft label correction by estimating the class-conditional label error probability. However, all of these methods overlook context relations, which can enhance adaptation performance without the need to access the source data.

% [width=1\textwidth,height=0.35\textheight]

\begin{figure}[!t]
\centering
\includegraphics[width=\textwidth]{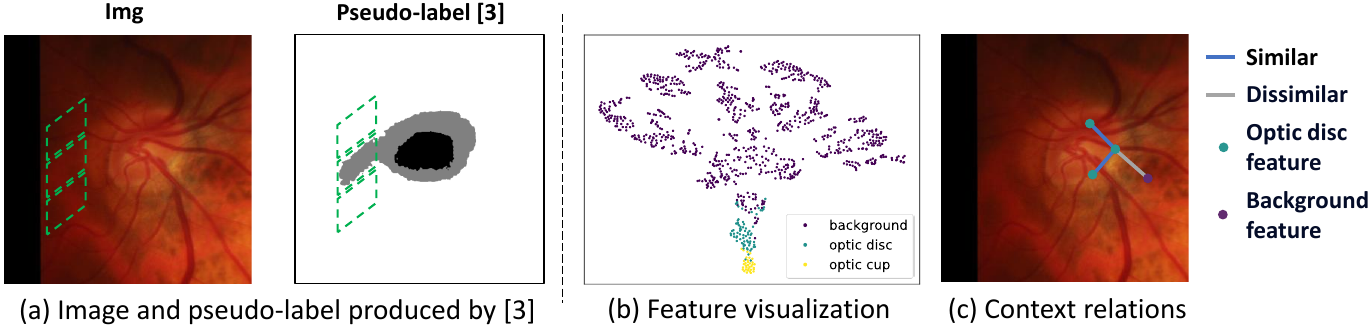}
\caption{(a) Example of context-inconsistent pseudo-labels. Due to domain gap, the pseudo-label of optic disc has irregular protuberance which is inconsistent with adjacent predictions. (b) t-SNE visualization of target pixel features produced by source model. Under domain shift, despite not aligning with source segmentor, target features of the same class still form a cluster. (c) Inspired by (b), context relations can be computed from feature distances.} \label{intro}
\end{figure}

We observe in our experiments (see Fig.~\ref{intro} (a)) that domain gaps can result in the source model making context-inconsistent predictions. For neighboring patches of an image with similar visual appearance, the source model can yield vastly different predictions. 
%recent work \cite{mic}
This phenomenon can be explained by the observation in \cite{shot} that target data shifts in the feature space, causing some data points to shift across the boundary of the source domain segmentor. 
%On the other hand, in a dense prediction task like segmentation, context information itself is a valuable source of semantic checking. These two aspects 
%Although the segmentation network has already modeled context relations to some extent, they do not capture context relations to the full potential without explicit learning. 
% The context-inconsistency motivates us to leverage context relations for refining the pseudo labels. 
% Additionally, it is found in \cite{neighborhood} and our experiments (see Fig.~\ref{intro}(b)) that target features produced by source model still form clusters, i.e., features of
% target data with the same class are close to each other. This inspires us to calculate context relations from feature distances (see Fig.~\ref{intro} (c)). Note that \cite{twoterms,neighborhood} deals with image-level features for classification while our method focuses on pixel-level features for segmentation.
The issue of context inconsistency motivates us to utilize context relations in refining pseudo-labels.
Moreover, it is observed in our experiments (as shown in Fig.~\ref{intro}(b)) that target features produced by the source model still form clusters, meaning that the features of target data points with the same class are closely located. This discovery led us to calculate context relations from feature distances; see Fig.~\ref{intro}(c). 
%Note that while \cite{twoterms,neighborhood} deal with image-level features for classification, our method focuses on pixel-level features for segmentation. \xmli{Confused about the last sentence. }

In this paper, we present a novel context-aware pseudo-label refinement (\textbf{CPR}) framework for source-free unsupervised domain adaptation. Firstly, we develop a context-similarity learning module, where context relations are computed from distances of features via a context-similarity head. This takes advantage of the intrinsic clustered feature distribution under domain shift \cite{twoterms,neighborhood}, where target features generated by the source encoder are close for the same class and faraway for different classes (see Fig.~\ref{intro} (b)). 
%\xmli{use xxx revision and xxx calibration to effectively xxxx. and xxx. Thirdly, We further develop a novel xxx which effectively.}
Secondly, context-aware revision is designed to leverage adjacent pseudo-labels for revising bad pseudo-labels, with aid of the learned context relations. Moreover, a calibration strategy is proposed, aiming to mitigate the negative effect brought about by the inaccurate learned context relations. Finally, the refined pseudo-labels are denoised with consideration of model knowledge and feature distribution \cite{dpl,weight} to select reliable pseudo-labels for domain adaptation. Experiments on cross-domain fundus image segmentation demonstrate our proposed framework outperforms the state-of-the-art source-free methods \cite{dpl,ud4r,generation}.

\begin{figure}[t]
\includegraphics[width=\textwidth]{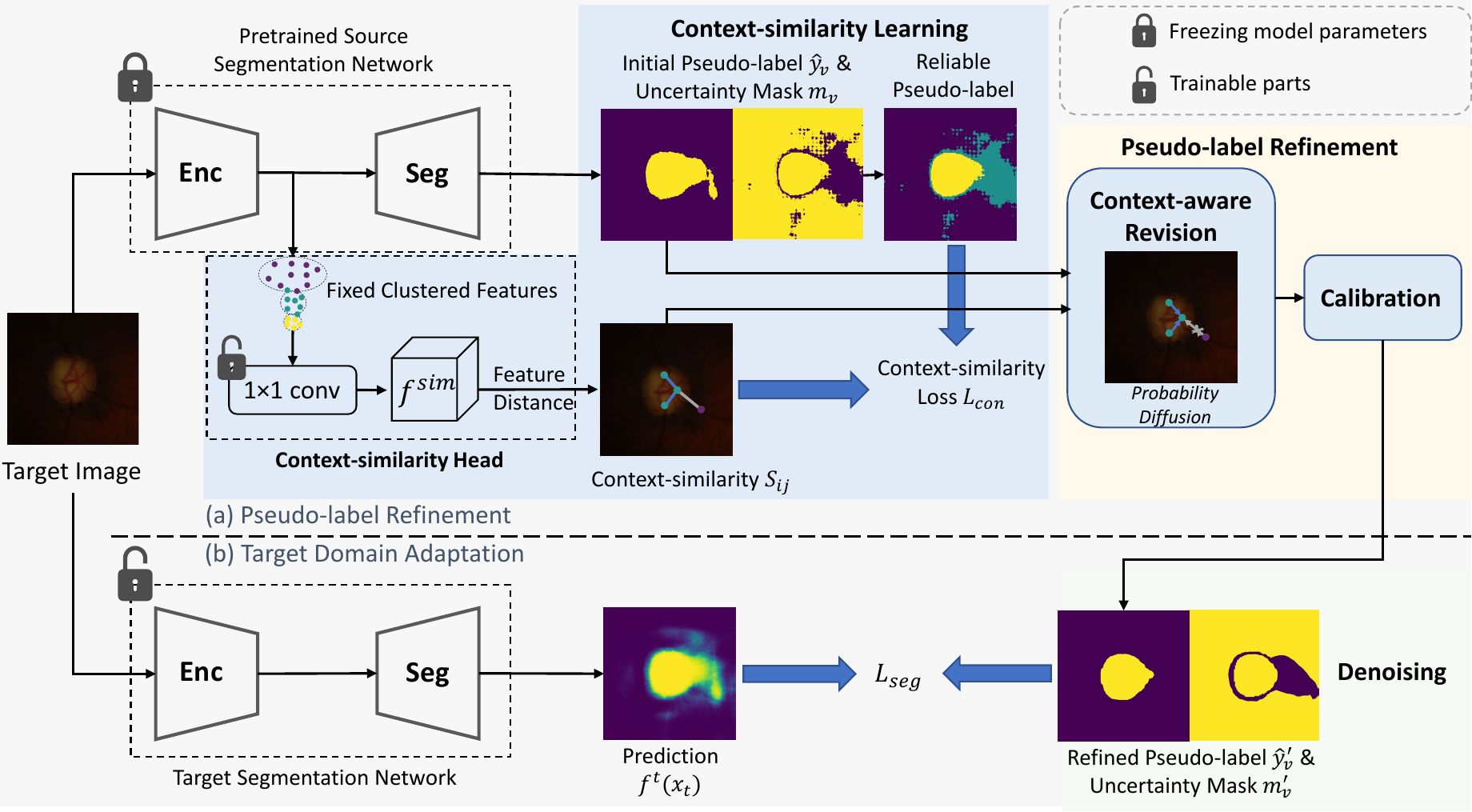}
\caption{Overview of the proposed context-aware pseudo-label refinement (CPR) framework for
SF-UDA. It consists of two stages: (a) The context-similarity head for computing context relations is trained by reliable pseudo-labels. The learned context similarities are then used to refine the pseudo-labels; (b) Only the refined pseudo-labels with high confidence supervise the training of the segmentation network. The network consists of a feature encoder (Enc) and a segmentor (Seg).} \label{method}
\end{figure}

\section{Method}
Fig.~\ref{method} illustrates our SF-UDA framework via context-aware pseudo-label refinement.
In this section, we first introduce the context-similarity learning scheme. Next, we propose the pseudo-label refinement strategy. Finally, we present the model training with the denoised refined pseudo-labels.

\subsection{Context-similarity learning}
In the SF-UDA problem, a source model $f^s: \mathcal{X}_s\rightarrow\mathcal{Y}_s$ is trained using the data $\{x_s^i,y_s^i\}_{i=1}^{n_s}$ from the source domain $\mathcal{D}_s=(\mathcal{X}_s,\mathcal{Y}_s)$, where $(x_s^i,y_s^i) \in (\mathcal{X}_s,\mathcal{Y}_s)$. $f^s$ is typically trained with a supervision loss of cross-entropy. Also an unlabeled dataset $\{x_t^i\}_{i=1}^{n_t}$ from the target domain $\mathcal{D}_t$ is given, where $x_t^i \in\mathcal{D}_t$. SF-UDA aims to learn a target model $f^t: \mathcal{X}_t\rightarrow\mathcal{Y}_t$ with only the source model $f^s$ and the target dataset $\{x_t^i\}_{i=1}^{n_t}$. In our fundus segmentation problem, $y^i \in \{0,1\}^{H\times W\times C}$, where $C$ is the number of classes and $C=2$ because there are two segmentation targets, namely optic cup and optic disc.

\vspace{0.5em}
\noindent\textbf{Architecture of context-similarity head.} %As found in \cite{neighborhood,twoterms},
Although the target features generated by source encoder do not align with the source segmentor, features of the same classes tend to be in the same cluster while those of different classes are faraway, as shown in Fig.~\ref{intro} (b). This indicates the source feature encoder is useful for computing context relations. Therefore, we freeze the source encoder and add an additional head to the encoder for learning context semantic relations, motivated by \cite{affinity}. A side benefit of freezing the source encoder is the training time and required memory can be reduced, as backward propagation is not needed on the encoder. Specifically, the feature map $f^{sim}$ is first obtained, where a $1\times 1$ convolution is applied for adaptation to the target task. Then the semantic similarity between coordinate $i$ and coordinate $j$ on the feature map is defined as
\begin{equation}
S_{ij}=\text{exp}\left\{-\Vert f^{sim}(x_i,y_i)-f^{sim}(x_j,y_j)\Vert_1 \right\}.
\end{equation}
Computing similarities between every pair of coordinates in a feature map is computationally costly. Thus, for each coordinate $i$, only similarities with coordinates $j$ lying within the circle of radius $r$ are considered in our implementation.

\vspace{0.5em}
\noindent\textbf{Training of context-similarity head.} Given a target image $x_t$, initial pseudo-labels and uncertainty mask can be obtained from the source model $f^s$ and $x_t$, following previous work \cite{dpl} as:
\begin{equation}\label{pseudo_label}
\begin{split}
p_{v,k}=&f^s(x_t)_v,k=1,\ldots,K,\\
p_v=\text{avg}(p_{v,1},&\ldots,p_{v,K}),u_v=\text{std}(p_{v,1},\ldots,p_{v,K}),\\
&\hat{y}_v=\mathbbm{1}[p_v\ge\gamma],
\end{split}
\end{equation}

\iffalse
\begin{equation}\label{mask}
\begin{split}
z^{\omega}\!=\!\frac{\sum\limits_vf_{l,v}\cdot\mathbbm{1}[\hat{y}_v=\omega]\mathbbm{1}[u_v<\eta]\cdot p_{v,\omega}}{\sum\limits_v\mathbbm{1}[\hat{y}_v=\omega]\mathbbm{1}[u_v<\eta]\cdot p_{v,\omega}},~&\omega\!\in\!\{\text{foreground \!(fg), \!background \!(bg)}\},\\
d^\omega_v=\Vert f_{l,v}-z^\omega&\Vert_2,\\
m_v=\mathbbm{1}[u_v<\eta](\mathbbm{1}[\hat{y}_v=1]\mathbbm{1}[d^{fg}_v<d^{bg}_v&]+\mathbbm{1}[\hat{y}_v=0]\mathbbm{1}[d^{fg}_v>d^{bg}_v]).
\end{split}
\end{equation}
\fi

\begin{align}\label{mask}
z^{\omega}\!=\!\frac{\sum\limits_vf_{l,v}\cdot\mathbbm{1}[\hat{y}_v=\omega]\mathbbm{1}[u_v<\eta]\cdot p_{v,\omega}}{\sum\limits_v\mathbbm{1}[\hat{y}_v=\omega]\mathbbm{1}[u_v<\eta]\cdot p_{v,\omega}},~&\omega\!\in\!\{\text{foreground \!(fg), \!background \!(bg)}\},\nonumber\\
d^\omega_v=\Vert f_{l,v}-z^\omega&\Vert_2,\\
m_v=\mathbbm{1}[u_v<\eta](\mathbbm{1}[\hat{y}_v=1]\mathbbm{1}[d^{fg}_v<d^{bg}_v&]+\mathbbm{1}[\hat{y}_v=0]\mathbbm{1}[d^{fg}_v>d^{bg}_v]).\nonumber
\end{align}

In Eq.~\ref{pseudo_label}, Monte Carlo Dropout \cite{mcdropout} is performed with $K$ forward passes through the source model, thereby calculating pseudo-label $\hat{y}_v$ and uncertainty $u_v$ for the $v$-th pixel. Eq.~\ref{mask} first extracts the class-wise prototypes $z^\omega$ from the feature map $f_{l,v}$ of the layer before the last convolution, then uncertainty mask $m_v$ is calculated by combining the distance to prototypes and uncertainty $u_v$. A pseudo-label for the $v$-th pixel is reliable if $m_v=1$.

Binary similarity label is then obtained. For two coordinates $i$ and $j$, similarity label $S^*_{ij}$ is $1$ if pseudo-labels $\hat{y}_i=\hat{y}_j$, and $0$ otherwise. Note only reliable pseudo-labels are considered to provide less noisy supervision.

The context-similarity head is trained with $S^*$. To address the class imbalance issue, the loss of each type of similarity (fg-fg, bg-bg, fg-bg) is calculated and aggregated \cite{affinity} as
\begin{equation}\label{similarity_loss}
\mathcal{L}_{con}=-\frac{1}{4}
\mathop{avg}_{\substack{\hat{y}_i=\hat{y}_j=1\\m_i=m_j=1}}(\text{log}S_{ij})-\frac{1}{4}
\mathop{avg}_{\substack{\hat{y}_i=\hat{y}_j=0\\m_i=m_j=1}}(\text{log}S_{ij})-\frac{1}{2}
\mathop{avg}_{\substack{\hat{y}_i\neq\hat{y}_j\\m_i=m_j=1}}(\text{log}(1-S_{ij})).
\end{equation}

\subsection{Context-similarity-based pseudo-label refinement}

%\vspace{0.5em}
\noindent\textbf{Context-aware revision.} The trained context-similarity head is utilized to refine the initial coarse pseudo-labels. Specifically, context-similarities $S_{ij}$ are computed by passing the target image through the source encoder and the trained head. Then the refined probability for the $i$-th coordinate is updated as the weighted average of the probabilities in a local circle around the $i$-th coordinate as
\begin{equation}\label{refine}
p_i^{re}=\sum_{\text{d}(i,j)\le r} \frac{{S_{ij}}^\beta}{\sum_{\text{d}(i,j)\le r}{S_{ij}}^\beta}\cdot p_j
\end{equation}
where $p_i^{re}$ is the revised probability and d($\cdot$) is the Euclidean distance. $\beta\ge 1$, in order to highlight the prominent similarities and ignore the smaller ones. By combining neighboring predictions based on context relations, revised probabilities are more robust. Eq.~\ref{refine} is performed iteratively for $t$ rounds, since revised probabilities can be used for further revision. 

\vspace{0.5em}
\noindent\textbf{Calibration.} The probability update by Eq.~\ref{refine} might be hurt by inaccurate context relations. We observe that for some classes (optic cup for fundus segmentation) with worse pseudo-labels, the context-similarity for ``fg-bg" is not learned well. Consequently, the probability of background incorrectly propagates to that of foreground, making the probability of foreground lower. To tackle this issue, the revised probability is calibrated as 
\begin{equation}\label{calibrate}
p_i'=\frac{p_i^{re}}{\mathop{max}_j(p_j^{re})}.
\end{equation}
The decreased probability is rectified by the maximum value in the image, considering the maximum probability (e.g., in the center of a region) after  calibration of a class should be close to $1$.

\subsection{Model adaptation with denoised pseudo labels}
The refined pseudo-labels can be obtained by $\hat{y}_v'=\mathbbm{1}[p_v'\ge\gamma]$. However, noisy pseudo-labels inevitably exist. The combination of model knowledge and target feature distribution shows the best estimation of sample confidence \cite{weight}. To this end, reliable pseudo-labels are selected at pixel-level and class-level \cite{dpl} as
\begin{equation}
\begin{split}
&m'_{v,p}=\mathbbm{1}(p_v'<\gamma_{low} \:\text{or} \: p_v'>\gamma_{high})\\
m'_{v,c}=\mathbbm{1}(\hat{y}'_v&=1)\mathbbm{1}(d^{fg}_v<d^{bg}_v)+\mathbbm{1}(\hat{y}'_v=0)\mathbbm{1}(d^{fg}_v>d^{bg}_v),
\end{split}
\end{equation}
in which $\gamma_{low}$ and $\gamma_{high}$ are two thresholds for filtering out pseudo-labels without confident probabilities. $d^{fg}_v$ and $d^{bg}_v$ are the distances to feature prototypes as computed in Eq.~\ref{mask}. The final label selection mask is the intersection of $m'_{v,p}$ and $m'_{v,c}$, i.e., $m_v'=m'_{v,p}\cdot m'_{v,c}$.
The target model is trained under the supervision of pseudo-labels selected by $m_v'$, with cross-entropy loss: 
\begin{equation}
\mathcal{L}_{seg}=-\sum_{v}m_v'\cdot \left[\hat{y}'_v\cdot {\rm log}(f^t(x_{t})_v)+(1-\hat{y}'_v)\cdot {\rm log}(1-f^t(x_{t})_v)\right].
\end{equation}
%To avoid the class-imbalance issue, focal loss is introduced \cite{focal}. Unlike dice loss \cite{dice_loss}, focal loss is compatible with the incomplete pseudo-labels.
\iffalse
\begin{equation}
\mathcal{L}_{seg}=-\sum_{v}m_v'\cdot \left[\hat{y}'_v\cdot (1-f^t(x_{t})_v)^2\cdot {\rm log}(f^t(x_{t})_v)+(1-\hat{y}'_v)\cdot (f^t(x_{t})_v)^2\cdot {\rm log}(1-f^t(x_{t})_v)\right].
\end{equation}
\fi

\section{Experiments}

\noindent\textbf{Datasets.}
For a fair comparison, we follow prior work~\cite{dpl} to select three mainstream datasets for fundus image segmentation,~\ie, Drishti-GS \cite{drishti_gs}, RIM-ONE-r3 \cite{rim_one}, and the validation set of REFUGE challenge \cite{refuge}. These datasets are split into $50/51$, $99/60$, and $320/80$ for training/testing, respectively. 
% In the first scenario, the Drishti-GS and RIM-ONEr3 datasets are used as the source and target domains, which presents a large domain gap.
% %
% In the second scenario, REFUGE is the source domain and Drishti-GS is the target domain, which shows a small domain gap.

\if
The proposed approach is validated on the optic cup and disc segmentation problem, with fundus image datasets collected at different clinical centers. Our experiments employ three datasets: Drishti-GS \cite{drishti_gs}, RIM-ONE-r3 \cite{rim_one} and the validation set of REFUGE challenge \cite{refuge}, which have training/testing sets of 50/51, 99/60 and 320/80 (following \cite{fundus_segmentation}) images, respectively. 
Experiments are performed in two domain adaptation scenarios to represent the cases of relatively large/small domain gaps. In the first scenario, the Drishti-GS dataset is used as the source domain, and the RIM-ONE-r3 dataset is used as the target domain. In the second scenario, REFUGE is the source domain and Drishti-GS is the target domain.
\fi

% We evaluate the segmentation performance with two widely used metrics: Dice coefficient and Average Surface Distance. 
\vspace{0.5em}
\noindent\textbf{Implementation details and evaluation metrics.}
% The training in the source domain follows previous work \cite{beal} without the boundary branch. 
Following prior works~\cite{dpl,beal,ud4r}, our segmentation network is MobileNetV2-adapted~\cite{mobilenetv2} DeepLabv3+~\cite{deeplabv3p}.
The context-similarity head comprises two branches for optic cup and optic disc, respectively. Each branch includes a $1\times 1$ convolution and a similarity feature map.
The threshold $\gamma$ for determining pseudo-labels is set to 0.75, referring to \cite{beal}. The radius $r$ in Eq.~\ref{refine}, the $\beta$ in Eq.~\ref{refine} and the iteration number $t$ are  set to $4$, $2$ and $4$ respectively. The two thresholds for filtering out unconfident refined pseudo-labels are empirically set as $\gamma_{low}=0.4$ and $\gamma_{high}=0.85$, respectively. Each image is pre-processed by clipping a $512\times 512$ optic disc region \cite{beal}. The same augmentations as in \cite{dpl,ud4r} are applied, including Gaussian noise, contrast adjustment, and random erasing.  The Adam optimizer is adopted with learning rates of 3e-2 and 3e-4 in the context-similarity learning stage and the target domain adaptation stage respectively. The momentum of the Adam optimizer is set to $0.9$ and $0.99$. The batch size is set to $8$. The context-similarity head is trained for $16$ epochs and the target model is trained for $10$ epochs. The implementation is carried out via PyTorch on a single NVIDIA GeForce RTX 3090 GPU. For evaluation, we adopt the widely used Dice coefficient and Average Surface Distance (ASD).

\if
The training in the source domain follows previous work \cite{beal} without the boundary branch. Following \cite{dpl,ud4r}, we use the MobileNetV2 \cite{mobilenetv2} adapted DeepLabv3+ \cite{deeplabv3p} as the network backbone. The context-similarity head has two branches for optic cup and disc respectively, with each branch having a $1\times 1$ convolution and a similarity feature map. The head is trained for $16$ epochs. The radius $r$ when computing context similarities is set as $4$. We set $\beta$ in Eq.~\ref{refine} to 2 and the iteration number $t$ to 4. For Eq.~\ref{pseudo_label} and Eq.~\ref{mask}, 100 forward passes are conducted, with a dropout rate of 0.5, and the uncertainty threshold $\eta$ is 0.05, following \cite{dpl}. The threshold $\gamma$ for determining pseudo-labels is 0.75 referring to \cite{beal}. The two thresholds $\gamma_{low}$ and $\gamma_{high}$ for filtering out unconfident refined pseudo-labels are empirically set as $0.4$ and $0.85$, respectively. Every image is resized to $512\times 512$ as the model input. The same augmentations as in \cite{dpl,ud4r} are applied, including Gaussian noise, contrast
adjustment, and random erasing. Focal loss \cite{focal} is introduced to make the training process more stable. We adopt Adam optimizer with learning rate of 3e-4, and momentum of 0.9 and 0.99. We set the batch size to 8 and train for $10$ epochs. The model is implemented via Pytorch 1.12.1 on one NVIDIA GeForce RTX 3090 GPU. For evaluation, we adopt the widely used Dice coefficient (the higher, the better) and Average Surface Distance (ASD; the lower, the better). 
\fi

\begin{figure}[!t]
\includegraphics[width=\textwidth]{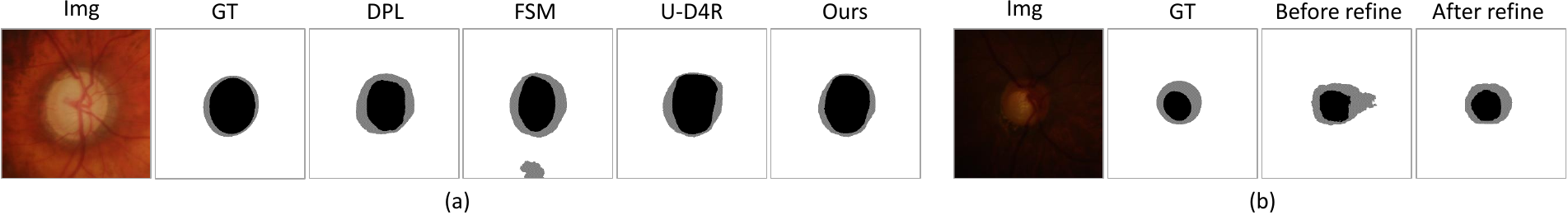}
\caption{On the Drishti-GS to RIM-ONE-r3 adaptation: (a) Qualitative comparison of the optic cup and disc segmentation results with different methods. (b) An example of pseudo-label change with the proposed refinement scheme.} \label{qualitative}
\end{figure}

\vspace{0.5em}
\noindent\textbf{Comparison with state-of-the-arts.} 
Table~\ref{quanti_result} shows the comparison of our method with the state-of-the-art SF-UDA methods.
Besides three SOTA methods,~\ie,  DPL \cite{dpl}, FSM \cite{generation}, and U-D4R \cite{ud4r}, we also report the adaptation result without adaptation and the result with fully supervised learning (denoted as ``upper bound").
The results show that our approach achieves clear improvements over the previous methods, %~\eg, 
owing to the proposed pseudo-label refinement scheme which takes advantage of the feature distribution property under domain shift to learn context relations and utilizes valuable context information to rectify pseudo-labels. Fig.~\ref{qualitative} (a) shows a qualitative comparison.
\if
In Table~\ref{quanti_result}, our method is compared with the state-of-the-arts of SF-UDA, i.e., DPL \cite{dpl}, FSM \cite{generation} and U-D4R \cite{ud4r}. Besides, the adaptation result without adaptation and the result with fully supervised learning (denoted as ``upper bound") are also included. As observed, DPL \cite{dpl} has small improvements over the baseline, because the model's discrimination ability is harmed by the coarse pseudo-labels  generated from Monte Carlo Dropout \cite{mcdropout}. FSM \cite{generation} performs well in the first adaptation scenario but worse in the second. The reason might be when the source model already has high performance, it requires very accurate source-like images to further boost performance. U-D4R \cite{ud4r} presents competitive results, but is still sub-optimal. Our approach achieves clear improvements over the previous methods, owing to the proposed pseudo-label refinement scheme which takes advantage of the feature distribution property under domain shift to learn context relations and utilizes valuable context information to rectify pseudo-labels. Fig.~\ref{qualitative} (a) shows a qualitative comparison.
\fi

\begin{table}[!t]
\centering
\caption{Comparison with state-of-the-arts on two settings. ``W/o adaptation'' refers to directly evaluating the source model on the target dataset. ``Upper bound'' refers to training the model on the target dataset with labels.}\label{quanti_result}
\begin{tabular}{c|c|c|c|c|c|c}

\toprule[1.5pt]

\multirow{2}{*}{Methods} & \multicolumn{3}{c|}{Dice[\%]$\uparrow$} & \multicolumn{3}{c}{ASD[pixel]$\downarrow$}\\\cline{2-7}
& Optic cup & Optic disc & Avg & Optic cup & Optic disc & Avg \\
\hline
\multicolumn{7}{c}{Source: Drishti-GS; Target: RIM-ONE-r3}\\\hline
W/o adaptation& 70.84 & 89.94 & 80.39 & 13.44 & 10.76 & 12.10\\
Upper bound & 83.81 & 96.61 & 90.21 & 6.92 & 2.96 & 4.94\\\hline
DPL \cite{dpl} & 71.70 & 92.52 & 82.11 & 12.49 & 7.34 & 9.92\\
FSM \cite{generation} & 74.34 & 91.41 & 82.88 & 14.52 & 10.30 & 12.41\\
U-D4R \cite{ud4r} & 73.48 & 93.18 & 83.33 & 10.18 & 6.15 & 8.16\\
CPR (ours) & \textbf{75.02} & \textbf{95.03} & \textbf{85.03} & \textbf{9.84} & \textbf{4.32} & \textbf{7.08}\\\hline

\multicolumn{7}{c}{Source: REFUGE; Target: Drishti-GS}\\\hline
W/o adaptation& 79.80 & 93.89 & 86.84 & 13.25 & 6.70 & 9.97\\
Upper bound & 89.63 & 96.80 & 93.22 & 6.65 & 3.55 & 5.10\\\hline
DPL \cite{dpl} & 82.04 & 95.27 & 88.65 & 12.14 & 5.32 & 8.73\\
FSM \cite{generation} & 79.30 & 94.34 & 86.82 & 13.79 & 5.95 & 9.87\\
U-D4R \cite{ud4r} & 81.82 & 95.98 & 88.90 & 12.21 & 4.45 & 8.33\\
CPR (ours) & \textbf{84.49} & \textbf{96.16} & \textbf{90.32} & \textbf{10.19} & \textbf{4.23} & \textbf{7.21}\\

\bottomrule[1.5pt]

\end{tabular}
\end{table}

%\xmli{Ablation study is not clear. We should clearly demonstrate the effects of different components, such as having several para}

%\subsubsection{Ablation study on different modules.}
\vspace{0.5em}
\noindent\textbf{Ablation study on different modules.} Table.~\ref{ablation} provides a quantitative analysis to investigate the function of each module. Each component shows its importance in improving the adaptation performance. Particularly, without our pseudo-label refinement, an obvious decrease of segmentation performance can be witnessed, revealing its necessity. Without calibration, the segmentation performance degrades significantly, which is because the probabilities without calibration do not have correct absolute values. This demonstrates calibration is a necessary step after the revision. Denoising filters out unreliable pseudo-labels by taking into account individual probabilities and feature distribution, thus providing more correct guidance. Integrating all the components completes our framework and yields the best result.

\iffalse
\begin{table}
\centering
\caption{Quantitative ablation study.}\label{ablation}
\begin{tabular}{|c|c|c|c|}
\hline
\multirow{2}{*}{Methods} & \multicolumn{3}{c|}{Dice[\%]}\\\cline{2-4}
& Optic cup & Optic disc & Avg \\\hline
ours & \textbf{75.03} & \textbf{95.02} & \textbf{85.02}\\\hline
w/o denoising & 74.52 & 92.69 &83.61\\\hline
w/o refinement & 72.34 & 94.13 & 83.23\\\hline
w/o calibration & 69.80 & 94.95 & 82.38\\\hline

\end{tabular}
\end{table}
\fi

\begin{table}[t]
\centering
\caption{Quantitative ablation study on the Drishti-GS to RIM-ONE-r3 adaptation.}\label{ablation}
\begin{tabular}{c|c|c|c|c|c}
\toprule[1.5pt]
\multicolumn{2}{c|}{Pseudo-label refinement} & \multirow{2}{*}{Denoising} & \multicolumn{3}{c}{Dice[\%]}\\\cline{1-2}\cline{4-6}
Context-aware revision & Calibration & & Optic cup & Optic disc & Avg \\\hline
 \XSolidBrush & \XSolidBrush & \XSolidBrush & 67.25 & 93.73 & 80.49\\\hline
 \Checkmark & \XSolidBrush & \XSolidBrush & 53.73 & 92.67 & 73.20 \\\hline
\Checkmark & \XSolidBrush & \Checkmark & 69.80 & 94.95 & 82.38 \\\hline
\Checkmark & \Checkmark & \XSolidBrush & 74.68 & 93.10 & 83.89 \\\hline
 \XSolidBrush & \XSolidBrush & \Checkmark & 72.34 & 94.13 & 83.23 \\\hline
\Checkmark & \Checkmark & \Checkmark & \textbf{75.02} & \textbf{95.03} & \textbf{85.03} \\ 
\bottomrule[1.5pt]
\end{tabular}
\end{table}

\noindent \textbf{Ablation study on pseudo-label refinement.}
Ablation study is conducted to verify the effectiveness of the pseudo-label refinement strategy. As shown in Table.~\ref{pseudo_com}, after refinement, the quality of the pseudo-label is clearly promoted, leading to more accurate supervision for target domain adaptation. For the pseudo-label of optic disc which originally has high accuracy, our refinement scheme encouragingly achieves a boost of $3.5\%$, showing the robustness of our refinement scheme for different quality of initial pseudo-labels. Without calibration, the accuracy of the pseudo-label of optic cup is substantially dropped, indicating it is an indispensable part of the overall scheme. Fig.~\ref{qualitative} (b) visualizes an example of the evolution of the pseudo-label. As can be seen, the context-inconsistent region is clearly improved.

\begin{table}
\centering
\caption{Comparison of pseudo-label quality of the training set with different methods on the Drishti-GS to RIM-ONE-r3 adaptation.}\label{pseudo_com}
\begin{tabular}{c|c|c}
\toprule[1.5pt]
\multirow{2}{*}{Methods} & \multicolumn{2}{c}{Dice[\%]}\\\cline{2-3}
& Optic cup & Optic disc \\\hline
Initial pseudo-label \cite{dpl} & 67.66 & 90.01\\\hline
Refined pseudo-label & \textbf{72.01} & \textbf{93.51}\\\hline
Refined pseudo-label (w/o calibration) & 58.34 & 93.40  \\
\bottomrule[1.5pt]
\end{tabular}
\end{table}

\iffalse
\begin{figure}
\centering
\includegraphics[scale=0.4]{pl_change.png}
\caption{Examples of pseudo-label change with the proposed refinement scheme.} \label{pl_change}
\end{figure}
\fi

\section{Conclusion}
This work presents a novel SF-UDA method for the fundus image segmentation problem. We propose to explicitly learn context semantic relations to refine pseudo-labels. Calibration is performed to compensate for the wrong revision caused by inaccurate context relations.  The performance is further boosted via the denoising scheme, which provides reliable guidance for adaptation. Our experiments on cross-domain fundus image segmentation show that our method outperforms the state-of-the-art SF-UDA approaches.

%\noindent\textbf
\subsubsection{Acknowledgement.}
This work was partially supported by the Hong Kong Innovation and Technology Fund under Project ITS/030/21, as well as by the HKUST-BICI Exploratory Fund (HCIC-004) and Foshan HKUST Projects under Grants FSUST21-HKUST10E and FSUST21-HKUST11E.

\iffalse
\begin{figure}
\includegraphics[width=\textwidth]{fig1.eps}
\caption{A figure caption is always placed below the illustration.
Please note that short captions are centered, while long ones are
justified by the macro package automatically.} \label{fig1}
\end{figure}

\begin{theorem}
This is a sample theorem. The run-in heading is set in bold, while
the following text appears in italics. Definitions, lemmas,
propositions, and corollaries are styled the same way.
\end{theorem}
%
% the environments 'definition', 'lemma', 'proposition', 'corollary',
% 'remark', and 'example' are defined in the LLNCS documentclass as well.
%
\begin{proof}
Proofs, examples, and remarks have the initial word in italics,
while the following text appears in normal font.
\end{proof}
\fi

\bibliographystyle{splncs04}
\bibliography{paper}

\begin{thebibliography}{10}
\providecommand{\url}[1]{\texttt{#1}}
\providecommand{\urlprefix}{URL }
\providecommand{\doi}[1]{https://doi.org/#1}

\bibitem{affinity}
Ahn, J., Kwak, S.: Learning pixel-level semantic affinity with image-level
  supervision for weakly supervised semantic segmentation. In: Proceedings of
  the IEEE conference on computer vision and pattern recognition. pp.
  4981--4990 (2018)

\bibitem{adaent}
Bateson, M., Kervadec, H., Dolz, J., Lombaert, H., Ben~Ayed, I.: Source-relaxed
  domain adaptation for image segmentation. In: International Conference on
  Medical Image Computing and Computer-Assisted Intervention. pp. 490--499.
  Springer (2020)

\bibitem{dpl}
Chen, C., Liu, Q., Jin, Y., Dou, Q., Heng, P.A.: Source-free domain adaptive
  fundus image segmentation with denoised pseudo-labeling. In: International
  Conference on Medical Image Computing and Computer-Assisted Intervention. pp.
  225--235. Springer (2021)

\bibitem{deeplabv3p}
Chen, L.C., Zhu, Y., Papandreou, G., Schroff, F., Adam, H.: Encoder-decoder
  with atrous separable convolution for semantic image segmentation. In:
  Proceedings of the European conference on computer vision (ECCV). pp.
  801--818 (2018)

\bibitem{de}
Ding, N., Xu, Y., Tang, Y., Xu, C., Wang, Y., Tao, D.: Source-free domain
  adaptation via distribution estimation. In: Proceedings of the IEEE/CVF
  Conference on Computer Vision and Pattern Recognition. pp. 7212--7222 (2022)

\bibitem{fundus_seg_background}
Fu, H., Cheng, J., Xu, Y., Wong, D.W.K., Liu, J., Cao, X.: Joint optic disc and
  cup segmentation based on multi-label deep network and polar transformation.
  IEEE transactions on medical imaging  \textbf{37}(7),  1597--1605 (2018)

\bibitem{rim_one}
Fumero, F., Alay{\'o}n, S., Sanchez, J.L., Sigut, J., Gonzalez-Hernandez, M.:
  Rim-one: An open retinal image database for optic nerve evaluation. In: 2011
  24th international symposium on computer-based medical systems (CBMS).
  pp.~1--6. IEEE (2011)

\bibitem{mcdropout}
Gal, Y., Ghahramani, Z.: Dropout as a bayesian approximation: Representing
  model uncertainty in deep learning. In: international conference on machine
  learning. pp. 1050--1059. PMLR (2016)

\bibitem{prompt}
Hu, S., Liao, Z., Xia, Y.: Prosfda: Prompt learning based source-free domain
  adaptation for medical image segmentation. arXiv preprint arXiv:2211.11514
  (2022)

\bibitem{perturb}
Jing, M., Zhen, X., Li, J., Snoek, C.G.M.: Variational model perturbation for
  source-free domain adaptation. In: Oh, A.H., Agarwal, A., Belgrave, D., Cho,
  K. (eds.) Advances in Neural Information Processing Systems (2022),
  \url{https://openreview.net/forum?id=yTJze\_xm-u6}

\bibitem{uda}
Kamnitsas, K., Baumgartner, C., Ledig, C., Newcombe, V., Simpson, J., Kane, A.,
  Menon, D., Nori, A., Criminisi, A., Rueckert, D., et~al.: Unsupervised domain
  adaptation in brain lesion segmentation with adversarial networks. In:
  Information Processing in Medical Imaging: 25th International Conference,
  IPMI 2017, Boone, NC, USA, June 25-30, 2017, Proceedings 25. pp. 597--609.
  Springer (2017)

\bibitem{pseudolabel}
Lee, D.H., et~al.: Pseudo-label: The simple and efficient semi-supervised
  learning method for deep neural networks. In: Workshop on challenges in
  representation learning, ICML. vol.~3, p.~896. Atlanta (2013)

\bibitem{weight}
Lee, J., Jung, D., Yim, J., Yoon, S.: Confidence score for source-free
  unsupervised domain adaptation. In: International Conference on Machine
  Learning. pp. 12365--12377. PMLR (2022)

\bibitem{3cgan}
Li, R., Jiao, Q., Cao, W., Wong, H.S., Wu, S.: Model adaptation: Unsupervised
  domain adaptation without source data. In: Proceedings of the IEEE/CVF
  Conference on Computer Vision and Pattern Recognition. pp. 9641--9650 (2020)

\bibitem{shot}
Liang, J., Hu, D., Feng, J.: Do we really need to access the source data?
  source hypothesis transfer for unsupervised domain adaptation. In:
  International Conference on Machine Learning. pp. 6028--6039. PMLR (2020)

\bibitem{memory}
Liu, X., Xing, F., El~Fakhri, G., Woo, J.: Memory consistent unsupervised
  off-the-shelf model adaptation for source-relaxed medical image segmentation.
  Medical Image Analysis  \textbf{83},  102641 (2022)

\bibitem{bnmiccai}
Liu, X., Xing, F., Yang, C., El~Fakhri, G., Woo, J.: Adapting off-the-shelf
  source segmenter for target medical image segmentation. In: International
  Conference on Medical Image Computing and Computer-Assisted Intervention. pp.
  549--559. Springer (2021)

\bibitem{refuge}
Orlando, J.I., Fu, H., Breda, J.B., Van~Keer, K., Bathula, D.R., Diaz-Pinto,
  A., Fang, R., Heng, P.A., Kim, J., Lee, J., et~al.: Refuge challenge: A
  unified framework for evaluating automated methods for glaucoma assessment
  from fundus photographs. Medical image analysis  \textbf{59},  101570 (2020)

\bibitem{ug}
Roy, S., Trapp, M., Pilzer, A., Kannala, J., Sebe, N., Ricci, E., Solin, A.:
  Uncertainty-guided source-free domain adaptation. In: European Conference on
  Computer Vision. pp. 537--555. Springer (2022)

\bibitem{prabhu}
S, P.T., Fleuret, F.: Uncertainty reduction for model adaptation in semantic
  segmentation. In: Proceedings of the IEEE/CVF Conference on Computer Vision
  and Pattern Recognition (CVPR). pp. 9613--9623 (June 2021)

\bibitem{mobilenetv2}
Sandler, M., Howard, A., Zhu, M., Zhmoginov, A., Chen, L.C.: Mobilenetv2:
  Inverted residuals and linear bottlenecks. In: Proceedings of the IEEE
  conference on computer vision and pattern recognition. pp. 4510--4520 (2018)

\bibitem{drishti_gs}
Sivaswamy, J., Krishnadas, S., Chakravarty, A., Joshi, G., Tabish, A.S.,
  et~al.: A comprehensive retinal image dataset for the assessment of glaucoma
  from the optic nerve head analysis. JSM Biomedical Imaging Data Papers
  \textbf{2}(1), ~1004 (2015)

\bibitem{tent}
Wang, D., Shelhamer, E., Liu, S., Olshausen, B., Darrell, T.: Tent: Fully
  test-time adaptation by entropy minimization. In: International Conference on
  Learning Representations (2021),
  \url{https://openreview.net/forum?id=uXl3bZLkr3c}

\bibitem{beal}
Wang, S., Yu, L., Li, K., Yang, X., Fu, C.W., Heng, P.A.: Boundary and
  entropy-driven adversarial learning for fundus image segmentation. In:
  Medical Image Computing and Computer Assisted Intervention--MICCAI 2019: 22nd
  International Conference, Shenzhen, China, October 13--17, 2019, Proceedings,
  Part I 22. pp. 102--110. Springer (2019)

\bibitem{ud4r}
Xu, Z., Lu, D., Wang, Y., Luo, J., Wei, D., Zheng, Y., Tong, R.K.y.: Denoising
  for relaxing: Unsupervised domain adaptive fundus image segmentation without
  source data. In: International Conference on Medical Image Computing and
  Computer-Assisted Intervention. pp. 214--224. Springer (2022)

\bibitem{generation}
Yang, C., Guo, X., Chen, Z., Yuan, Y.: Source free domain adaptation for
  medical image segmentation with fourier style mining. Medical Image Analysis
  \textbf{79},  102457 (2022)

\bibitem{twoterms}
Yang, S., Wang, Y., Wang, K., JUI, S., van~de weijer, J.: Attracting and
  dispersing: A simple approach for source-free domain adaptation. In: Oh,
  A.H., Agarwal, A., Belgrave, D., Cho, K. (eds.) Advances in Neural
  Information Processing Systems (2022),
  \url{https://openreview.net/forum?id=ZlCpRiZN7n}

\bibitem{neighborhood}
Yang, S., van~de Weijer, J., Herranz, L., Jui, S., et~al.: Exploiting the
  intrinsic neighborhood structure for source-free domain adaptation. Advances
  in Neural Information Processing Systems  \textbf{34},  29393--29405 (2021)

\bibitem{yao2022enhancing}
Yao, H., Hu, X., Li, X.: Enhancing pseudo label quality for semi-supervised
  domain-generalized medical image segmentation. In: Proceedings of the AAAI
  Conference on Artificial Intelligence. vol.~36, pp. 3099--3107 (2022)

\end{thebibliography}

\end{document}